\documentclass[manuscript]{acmart}

\usepackage{graphicx}
\usepackage{caption,subcaption}
\usepackage{microtype}
\usepackage{color,soul}
\usepackage{multirow}

\AtBeginDocument{%
  \providecommand\BibTeX{{%
    \normalfont B\kern-0.5em{\scshape i\kern-0.25em b}\kern-0.8em\TeX}}}

\setcopyright{acmcopyright}

\copyrightyear{2022}
\acmYear{2022}
\setcopyright{acmcopyright}\acmConference[UMAP '22]{Proceedings of the 30th ACM Conference on User Modeling, Adaptation and Personalization}{July 4--7, 2022}{Barcelona, Spain}
\acmBooktitle{Proceedings of the 30th ACM Conference on User Modeling, Adaptation and Personalization (UMAP '22), July 4--7, 2022, Barcelona, Spain}
\acmPrice{15.00}
\acmDOI{10.1145/3503252.3531304}
\acmISBN{978-1-4503-9207-5/22/07}

\begin{document}

\title[Is More Always Better?] {Is More Always Better? The Effects of Personal Characteristics and Level of Detail on the Perception of Explanations in a Recommender System}

\author{Mohamed Amine Chatti}
\affiliation{%
	\institution{University of Duisburg-Essen}
	  \city{Duisburg}
	\country{Germany}}
\email{mohamed.chatti@uni-due.de}

\author[1]{Mouadh Guesmi}
\affiliation{%
  \institution{University of Duisburg-Essen}
%  \streetaddress{1 Th{\o}rv{\"a}ld Circle}
  \city{Duisburg}
  \country{Germany}}
\email{mouadh.guesmi@stud.uni.de}

\author[1]{Laura Vorgerd}
\affiliation{%
	\institution{University of Duisburg-Essen}
	\city{Duisburg}
	\country{Germany}}
	\email{laura.vorgerd@stud.uni-due.de}
	
\author[1]{Thao Ngo}
\affiliation{%
	\institution{University of Duisburg-Essen}
	\city{Duisburg}
	\country{Germany}}
	\email{thao.ngo@uni-due.de}
	
\author[1]{Shoeb Joarder}
\affiliation{%
	\institution{University of Duisburg-Essen}
	\city{Duisburg}
	\country{Germany}}
	\email{shoeb.joarder@uni-due.de}
	
\author[1]{Qurat Ul Ain}
\affiliation{%
  \institution{University of Duisburg-Essen}
%  \streetaddress{1 Th{\o}rv{\"a}ld Circle}
  \city{Duisburg}
  \country{Germany}}
\email{qurat.ain@stud.uni.de}

\author[2]{Arham Muslim}
\affiliation{%
	\institution{National University of Sciences and Technology}
	\country{Pakistan}}
	\email{arham.muslim@seecs.edu.pk}

\renewcommand{\shortauthors}{M.A. Chatti, et al.}

\begin{abstract}
Despite the acknowledgment that the perception of explanations may vary considerably between end-users, explainable recommender systems (RS) have traditionally followed a one-size-fits-all model, whereby the same explanation level of detail is provided to each user, without taking into consideration individual user’s context, i.e., goals and personal characteristics. To fill this research gap, we aim in this paper at a shift from a one-size-fits-all to a personalized approach to explainable recommendation by giving users agency in deciding which explanation they would like to see. We developed a transparent Recommendation and Interest Modeling Application (RIMA) that provides on-demand personalized explanations of the recommendations, with three levels of detail (basic, intermediate, advanced) to meet the demands of different types of end-users. We conducted a within-subject study (N=31) to investigate the relationship between user’s personal characteristics and the explanation level of detail, and the effects of these two variables on the perception of the explainable RS with regard to different explanation goals. Our results show that the perception of explainable RS with different levels of detail is affected to different degrees by the explanation goal and user type. Consequently, we suggested some theoretical and design guidelines to support the systematic design of explanatory interfaces in RS tailored to the user’s context.
\end{abstract}


\begin{CCSXML}
<ccs2012>
   <concept>
       <concept_id>10003120.10003121.10003129</concept_id>
       <concept_desc>Human-centered computing~Interactive systems and tools</concept_desc>
       <concept_significance>500</concept_significance>
       </concept>
   <concept>
       <concept_id>10010147.10010178</concept_id>
       <concept_desc>Computing methodologies~Artificial intelligence</concept_desc>
       <concept_significance>500</concept_significance>
       </concept>
 </ccs2012>
\end{CCSXML}

\ccsdesc[500]{Human-centered computing~Interactive systems and tools}
\ccsdesc[500]{Computing methodologies~Artificial intelligence}

%\keywords{datasets, neural networks, gaze detection, text tagging}
\keywords{intelligent explanation interfaces; recommender systems; explainable recommendation; personalized explanation; personal characteristics}

\maketitle

\section{Introduction}
The explainability of recommendation systems (RS) has attracted considerable attention in recent years. Explainable recommendation refers to personalized recommendation algorithms that not only provide the user with the recommendations, but also provide explanations to make the user aware of why such items are recommended \cite{zhang2018explainable}. An explanation seeks to answer questions, also called intelligibility queries or types, such as what, why, how, what-if, why-not, how-to, and what-else \cite{lim2009assessing, mohseni2018}. Research on explainable recommendation has been focused on different dimensions and design choices. In addition to the intelligibility types, these include (a) explanation goal (e.g., transparency, scrutability, trust, effectiveness, persuasiveness, efficiency, satisfaction), (b) explanation style (e.g., content-based, collaborative-based, social, hybrid), (c) explanation scope (i.e. input: user model, process: algorithm, output: recommended items), and (d) explanation format (e.g., textual, visual) \cite{ain2022framework, nunes2017, tintarev2015explaining, zhang2018explainable}. Another crucial design choice in explainable recommendation relates to the level of explanation detail that should be provided to the end-user. Different explanation design choices, such as explanation style, scope, format, and level of detail will be affected by the explanation goal and user type \cite{mohseni2018}. Users may not be interested in all the information that the explanation can produce \cite{miller2019explanation}. Different users have different needs for explanation and explanations may cause negative effects (e.g., high cognitive load, confusion, lack of trust) if they are difficult to understand \cite{zhao2019users, gedikli2014,yang2020visual,kulesza2015principles, kizilcec2016much}. Thus, it is important to provide explanations with enough details to allow users to build accurate mental models of how the RS operates without overwhelming them. 

The effect of individual user differences and human factors on behaviors with explainable RS has only been studied very recently \cite{kouki2019personalized, millecamp2019, hernandez2020explaining, szymanski2021visual}. These studies showed that personal characteristics may have an impact on the perception of explanations, and provided motivation for the selection of different explanation information (i.e., content) for different users, depending on their context, i.e., goals and personal characteristics. However, in terms of design choice (i.e., intelligibility type, explanation style, scope, format, or level of detail), the majority of current designs of explainable RS still follow a one-size-fits-all approach that does not attempt to identify and address the needs and preferences of different users. As design choices in explainable recommendation will be affected by the user’s context, one natural direction is the advancement of current explanation techniques to meet the demands of different types of end-users. Like the recommendations themselves, explanations should be personalized for different end-users \cite{millecamp2019}. Thus, explainable RS are expected to provide the right explanations for the right group of users \cite{mohseni2018}. This requires a shift from a one-size-fits-all to a personalized approach to explainable recommendation, tailored to the needs and preferences of different users.

In this paper, we are particularly interested in the explanation level of detail as an important design choice in explainable recommendation. Recognizing that it is generally insufficient to take the explanation level of detail and user’s personal characteristics separately, we conducted a user study where we investigated the dependencies between these two factors and their effects on the user perception of different explanation goals (transparency, scrutability, trust, effectiveness, persuasiveness, efficiency, satisfaction). As a result, we derived some design guidelines to be considered when designing explanations with different levels of detail that align with explanation goals and user’s personal characteristics.  

To conduct this study, we developed a transparent Recommendation and Interest Modeling Application (RIMA) that provides on-demand personalized explanations of the recommendations with three different levels of detail (basic, intermediate, advanced), in order to meet the needs and preferences of different users. 
The objective of the study was to answer the following research question: \textbf{How do personal characteristics impact user perceptions of the explanation level of detail in terms of different explanation goals?}

The results of our study show that (1) it is important to provide explanations with different levels of detail to meet the demands of different users, (2) explanations should be designed with respect to specific explanation goals and specific user types, and (3) explanation content should be tailored to user data.

The main contribution of this paper is twofold: first, we take personalized explanation in RS to the design choice level by providing on-demand explanations with different levels of detail. Second, we provide evidence for a dependency relation between explanation goal, user type, and explanation level of detail. 

This paper is organized as follows. We first outline the background for this research and discuss related work. We then present the implementation of the different explanations in RIMA. Afterwards, we describe the user study, present the results, and discuss the implications of our findings. Based on these findings, we then present some design guidelines for providing personalized explanations in RS. Finally, we point out limitations, summarize the work, and outline future research plans.

\section{Related work}
%In the following, we discuss related work on explainable recommendations with a focus on providing explanation with different levels of details and the effects of personal characteristics on the perception of explainable RS.
\subsection{Explanation with different levels of detail}
The explanation level of detail is an important factor in the design process of explainable RS. In this work, the level of detail refers to the amount of information exposed in an explanation. In the field of explainable AI (XAI) in general, \citet{mohseni2018} argue that different user groups will have other goals in mind while using XAI systems. 
%For example, while data and machine learning (ML) experts might prefer highly-detailed visual explanations to help them inspect, debug, and optimize ML models, lay-users do not expect fully detailed explanations from an intelligent system. Instead, systems with lay-users as target groups aim to enhance the user experience with the system through improving their understanding and trust. 
In the same direction, \citet{miller2019explanation} argue that providing the exact algorithm which generated the specific recommendation is not necessarily the best explanation. People tend not to judge the quality of explanations around their generation process, but instead around their usefulness. 

Besides the goals of the users, another vital aspect that will influence their understanding of explanations are their cognitive capabilities. Results of previous research on XAI showed that for specific users or user groups, the detailed explanation does not automatically result in higher trust and user satisfaction because the provision of additional explanations increases cognitive effort \cite{kizilcec2016much, kulesza2015principles,zhao2019users, yang2020visual}. 
%Only when users have enough time to process the information and enough ability to figure out the meaning of the information, a higher level of detail in the explanation will lead to a better understanding. But as soon as the amount of information is beyond the users’ comprehension, the explanation could lead to information overload and bring confusion. Without the understanding of how the system works, users may perceive the system as not transparent enough, which could, in turn, reduce the users’ trust in the system \cite{zhao2019users}. 
\citet{kulesza2015principles} outlined a set of principles for designing explanations to personalize interactive machine learning. These principles include "Be Sound", “Be Complete” and “Don’t Overwhelm” implying a tradeoff between the amount of information in an explanation and the level of perceived transparency, trust, and satisfaction users develop when interacting with the AI system. \textit{Soundness} means telling nothing but the truth. It refers to the explanation fidelity, i.e., “the extent to which each component of an explanation’s content is truthful in describing the underlying system”. Evaluating soundness requires to compare the explanation with the learning system’s mathematical model, “the more these explanations reflect the underlying model, the more sound the explanation is”. \textit{Completeness} means telling the whole truth. It refers to “the extent to which all of the underlying system is described by the explanation”. A complete explanation informs users about all the information the learning system had at its disposal and how it used that information. The authors suggest that one method for evaluating completeness is via Lim and Dey's intelligibility types (e.g., input, model, why, what if, certainty) \cite{lim2009assessing}, with more complete explanations including more of these intelligibility types \cite{kulesza2015principles}. In an earlier study, \citet{kulesza2013too} considered ways intelligent agents should explain themselves to end users, especially focusing on how the soundness and completeness of the explanations impacts the end users’ mental models. The authors found that increasing completeness helped participants’ mental models and their perception of the cost/benefit tradeoff of attending to the explanations. On the other hand, when soundness was very low, participants experienced more mental demand and lost trust in the explanations. The study shows that there is a need to provide explanations with enough soundness and completeness in order to help users build an accurate mental model of how the system works without overwhelming them. In another study, \citet{kizilcec2016much} investigated the effects of three levels of system transparency on trust in an algorithmic interface in the context of peer assessment and concluded that designing for trust requires balanced interface transparency, i.e., “not too little and not too much”. In their study, \citet{yang2020visual} also observed that participants’ understanding of visual explanation was correlated with their trust. The authors showed that different visual explanations lead to different levels of trust and may cause inappropriate trust if an explanation is difficult to understand. In summary, these studies suggest that (1) different users demand different levels of explanation information, (2) the right explanation level of detail depends on the user’s context, i.e., goals and personal characteristics, and (3) providing the inappropriate explanation level of detail may cause negative effects.

While increasingly popular in XAI research, providing explanation with different levels of detail remains rare in the literature on explainable recommendation. Only the work presented in \cite {millecamp2019} provided explanations with varying level of details. Drawing on the findings from their study, the authors suggested that (1) users should be able to choose whether or not they wish to see explanations and (2) explanation components should be flexible enough to present varying level of details depending on users’ preferences. Following these design guidelines, the authors developed a music RS that not only allows users to choose whether or not to see the explanations by using a "Why?" button but also to select the level of detail by clicking on a "More/Hide" button. 

%A critical question in the research of explainable recommendation is whether the relationship between the level of detail and transparency is a linear one. To answer this question, we need first to differentiate between objective transparency and user-perceived transparency. Objective transparency means that the RS reveals the underlying algorithm of the recommendations. However, the algorithm might be too complex to be described in a human-interpretable manner. Therefore, it might be more appropriate to provide “justifications” instead of “explanations”, which are often superficial and more user-oriented. On the other hand, user-perceived transparency is based on the users’ subjective opinion about how good the system is capable of explaining its recommendations. In general, it can be assumed that a higher level of explanation detail increases the system’s objective transparency but is also associated with a risk of reducing the user-perceived transparency, and that this risk depends on the user’s personal characteristics \cite{gedikli2014}.
\subsection{Effects of personal characteristics}
Recent studies on explainable recommendation showed that personal characteristics have an effect on the perception of explanations and that it is important to take personal characteristics into account when designing explanations \cite{millecamp2019,kouki2019personalized,hernandez2020explaining,szymanski2021visual}. These studies investigated the effect of human factors, such as Big Five traits, need for cognition, and visualization familiarity and confirmed that users with specific personal characteristics will perceive and interact in different ways with an explainable RS. In particular, prior research investigated the effects of personal characteristics on the perception of different explanation styles (e.g., user-based, item-based, content-based, social) \cite{kouki2019personalized} and different explanation formats (textual, visual) \cite{hernandez2020explaining,szymanski2021visual}. However, the effects of personal characteristics on the perception of explanation with different levels of detail are under-explored in explainable recommendation research. 
\section{RIMA}
We developed the transparent Recommendation and Interest Modeling Application (RIMA) with the goal of explaining the recommendations with varying level of details. RIMA is a content-based RS that produces content-based explanations. It follows a user-driven personalized explanation approach by providing explanations with different levels of detail and empowering users to steer the explanation process the way they see fit \cite{guesmi2021input,guesmi2021open}. The application provides on-demand explanations, that is, the users can decide whether or not to see the explanation and they can also choose which level of explanation detail they want to see \cite{guesmi2021demand}. In this work, we focus on recommending tweets and leveraging explanatory visualizations to provide insights into the recommendation process. 
%(see Figure \ref{fig1}). 

%\begin{figure}
%	\centering
%	\includegraphics[height=0.45\textwidth,width=0.67\textwidth]{img/Tweet-rec.png}
%	\caption{Recommendation Interface in RIMA.}
%	\label{fig1}
%\end{figure}

%\subsection{Explaining the interest model}

%\subsubsection{Interest model inference}
\subsection{Interest model inference}
 The interest models in RIMA are inferred from users’ publications. The application uses Semantic Scholar IDs provided by users to gather their publications and the Semantic Interest Modeling Toolkit (SIMT) presented in \cite{chatti2021simt} to infer users' interest models based on their publications. It applies unsupervised keyphrase extraction algorithms on the collected publications to generate keyphrase-based interests. In order to address semantic issues, Wikipedia is leveraged as a knowledge base to map the keyphrases to Wikipedia pages and generate Wikipedia-based interests.

\subsection{Recommendation generation}
The aim of this part of the application is to provide tweet recommendations based on the inferred interest model. For obtaining the candidate tweets, we use the Twitter API to fetch tweets that contain one or more user interests that are used as input for the recommendation. We then apply an unsupervised keyphrase extraction algorithm on the fetched tweets to extract keywords from the tweet text. In order to compare the similarity between the user interests and the candidate tweets, we use word embedding techniques to generate vector representations of the interest model and the tweets. After getting the two embedding representations (i.e., interest model embedding and tweet embedding), we calculate the cosine similarity between them in order to obtain a semantic similarity score. Tweets with a semantic similarity score above a threshold of 40 \% will then be displayed to the user.

%\subsubsection{Personalized explanation of the recommendation}
\subsection{Explanation design}
The current explanation design was mainly the result of several brainstorming sessions involving the authors and students from the local university. In order to systematically design the explanations at different levels of details, we built upon Kulesza et al.’s principles of explanatory debugging \cite{kulesza2015principles} and Lim and Dey’s schema of intelligibility types \cite{lim2009assessing}. The visual design of the explanations was inspired by popular visualizations used in the literature on explainable RS, such as heatmaps and node-link diagrams \cite{gedikli2014}. RIMA aims to provide explanation with different levels of detail by varying the explanation \textit{soundness} and \textit{completeness}. Soundly explaining the recommendations requires accurately explaining the different components of the underlying recommendation algorithm in a detailed manner. Providing a complete explanation requires to inform users about all the information the RS knows about the user and how it used that information. \citet{kulesza2015principles} suggest that a complete explanation should include Lim and Dey’s input (information the system is aware of), why (the reasons underlying a specific decision), and how (an overview of the system’s decision making process) intelligibility types \cite{lim2009assessing}. RIMA provides three layered explanations (i.e., basic, intermediate, advanced) that the users can choose from, depending on whether they want more or less information. The intermediate and advanced explanations are hidden by default, but users are able to view these explanations on demand. Varying the explanation level of detail is achieved through manipulating and combining three levels of soundness and completeness (i.e., low, medium, high). 
%The aim of explaining the recommendation in RIMA is to provide a \textit{justification} on why a specific recommendation was presented and to help end-users' \textit{understanding} of how the recommendation works. This can improve users' mental model of the underlying recommendation algorithm. Further, \textit{transparency} of the RS can improve user experience through better understanding of the recommendation output, thus improving user interaction, \textit{trust}, and \textit{satisfaction} with the system.
%The application provides an on-demand personalized explanation of the recommendations (output) with three different levels of detail (basic, intermediate, and advanced). 
\subsubsection{Basic explanation}
\begin{figure} [ht]
	\centering
	\begin{subfigure}[b]{0.4\textwidth}
		\includegraphics[height=0.45\textwidth,width=\textwidth]{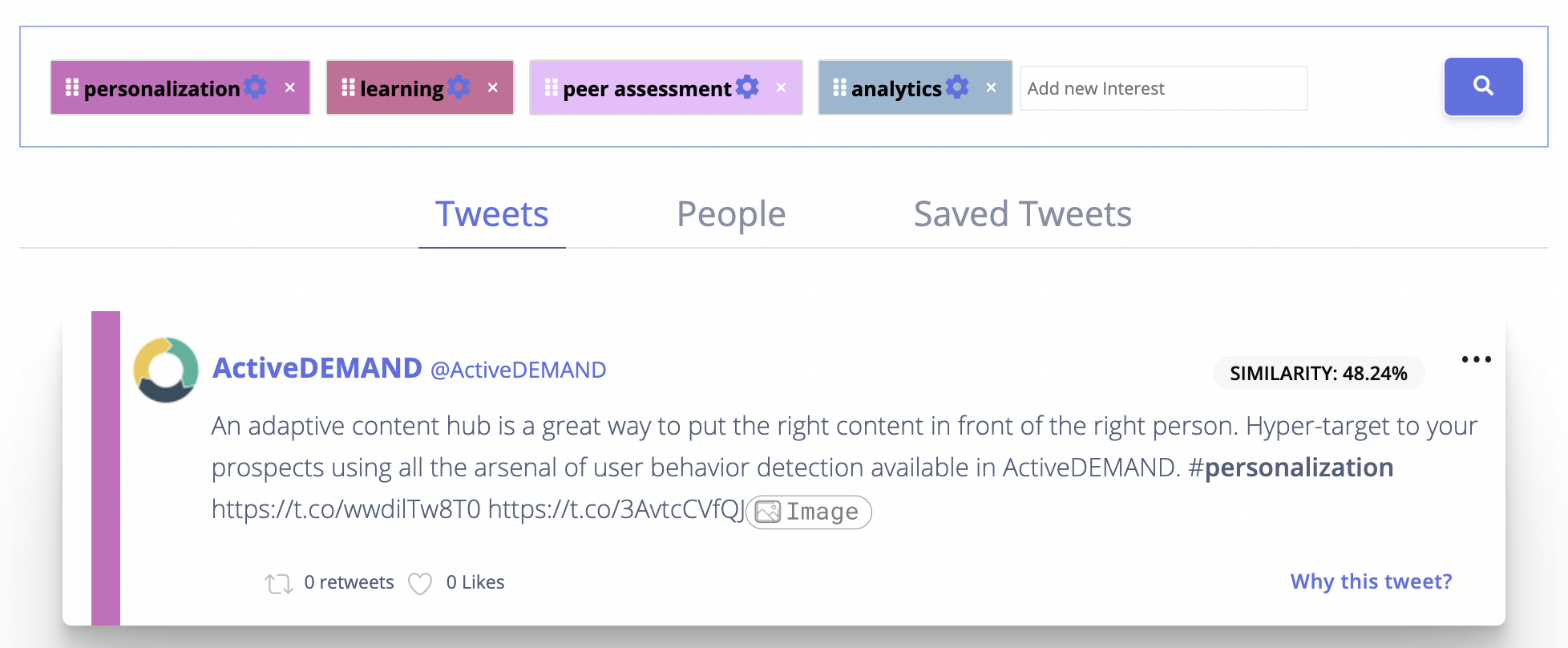}
		\caption{Basic explanation}
		\label{fig3:a}
	\end{subfigure}
	\hfill
	\begin{subfigure}[b]{0.53\textwidth}
	\includegraphics[height=0.4\textwidth, width=\textwidth]{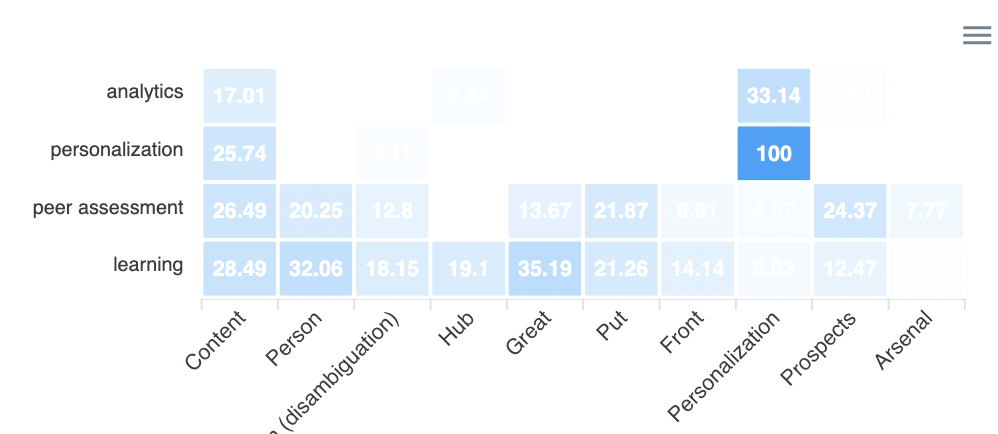}
		\caption{Intermediate explanation}
		\label{fig3:b}
	\end{subfigure}
	
	\hfill
	
	\begin{subfigure}[b]{0.7\textwidth}
		\includegraphics[width=\textwidth]{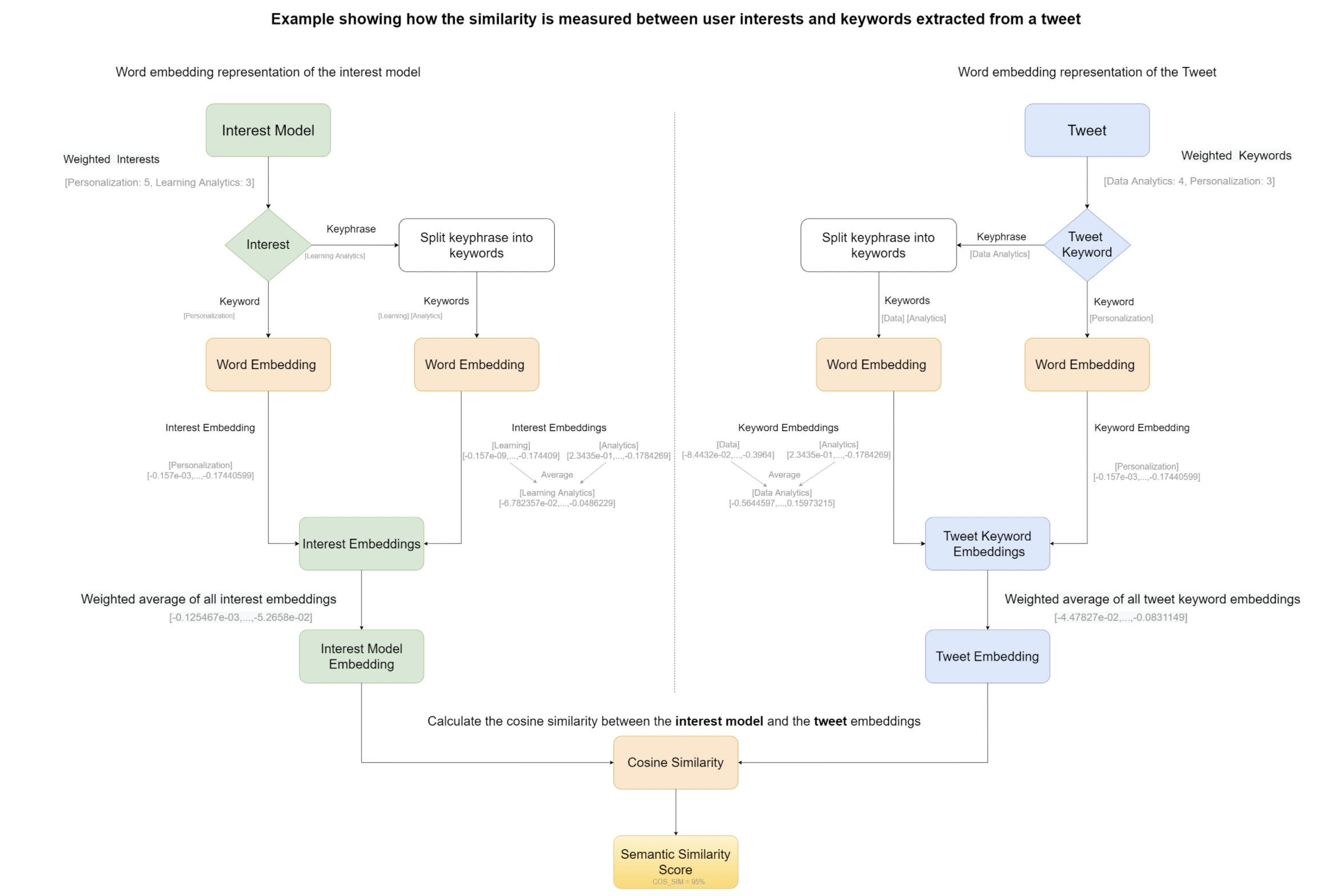}
		\caption{Advanced explanation}
		\label{fig3:c}
	\end{subfigure}
	\caption{Explaining the tweet recommendation with three levels of details.}
	\label{fig3}
\end{figure}

We designed the \textit{basic explanation} to explain the RS with \textit{low soundness} and \textit{medium completeness}. It provides a simple, broad, and low-fidelity explanation of the RS that does not provide concrete details about the underlying recommendation algorithm. In terms of completeness, this explanation includes \textit{input} and \textit{why} intelligibility types. It tells the user about the interest model available as input to the RS as well as the RS’ reasons for a specific recommendation, in an abstract manner. As shown in Figure \ref{fig3:a}, the search box is initially populated with the user’s top five interests, ordered by their weights as generated by the system. Users can also add new interests in the search box or remove existing ones. The system will use these interests as input for the recommendation process. The basic explanation is achieved using a color band to map the tweet to the related interest(s). Also, the interest will be highlighted in the text of the tweet to show that this tweet contains this specific word (interest). In addition to these two visual elements, we display the similarity score on the top right corner of the tweet to show the level of similarity between the user interests and the recommended tweet . The answer to the why intelligibilty type at this level is a visual representation of "because the tweet text contains your interest X and this tweet is Y\% similar to your interest profile".

%\begin{figure} [h]
%	\centering
%	\begin{subfigure}[b]{0.33\textwidth}
		%\includegraphics[height=0.4\textwidth,width=\textwidth]{img/TR basic explanation.png}
%		\caption{Basic explanation}
%		\label{fig3:a}
%	\end{subfigure}
%	\hfill
%	\begin{subfigure}[b]{0.33\textwidth}
%	\includegraphics[height=0.4\textwidth, width=\textwidth]{img/TR intermediate explanation.png}
%		\caption{Intermediate explanation}
%		\label{fig3:b}
%	\end{subfigure}
%	\hfill
%	\begin{subfigure}[b]{0.33\textwidth}
%		\includegraphics[width=\textwidth]{img/TR advanced explanation.png}
%		\caption{Advanced explanation}
%		\label{fig3:c}
%	\end{subfigure}
%	\caption{Explaining the tweet recommendation with three levels of details}
%	\label{fig3}
%\end{figure}

\subsubsection{Intermediate explanation}
For more details, the user can choose the \textit{intermediate explanation} level by clicking on "Why this tweet?" on the bottom right of the tweet. The intermediate explanation is an instantiation of an explanation with \textit{medium soundness} and \textit{medium completeness}. It provides a more concrete, mid-fidelity explanation of the underlying recommendation algorithm by showing the results of the computed similarities between user interests and keywords extracted from a recommended tweet. However, it does not accurately disclose the detailed steps of the recommendation process to users. In terms of completeness, only the \textit{input} and \textit{why} intelligibility types are included in this explanation. The intermediate explanation tells users about the sources of information available to the RS (i.e., user interests and tweets) and that the similarity between individual user interests and tweet keywords played a role in each recommendation. As shown in Figure \ref{fig3:b}, we used a Heatmap chart to show the semantic similarities between user interests and the keywords extracted from the text of the tweet. The x-axis represents the keywords extracted from the tweet and the y-axis represents the user's interests used in the recommendation. The cells show the computed semantic similarity scores between each interest and keyword. At this level, the answer to the why intelligibilty type is a visual representation of the similarities between all user interests used as input for the recommendation and all the keywords extracted from the tweet. 
\subsubsection{Advanced explanation}
To move to the \textit{advanced explanation} level, the user has to click on the "more" button on the bottom right of the intermediate explanation window. The advanced explanation further increases soundness and completeness. This explanation aims to achieve \textit{high soundness} by providing a high-fidelity explanation of the underlying recommendation process, as well as accurately detailing all the algorithmic steps used to compute the similarity between user interests and keywords extracted from a recommended tweet. To help ensure \textit{high completeness}, we exposed more information by adding the \textit{how} intelligibility type, in addition to the \textit{input} and \textit{why} intelligibility types. Following an explanation by example approach, the advanced explanation tells that the RS algorithm uses word embedding representations of a user’s interest model and a tweet as inputs to calculate a semantic similarity score between interest model and tweet embeddings. A tweet with a high similarity score will then be recommended (see Figure \ref{fig3:c}).
%The aim of the advanced explanation is to explain "how" the recommendation algorithm works. This is achieved by following an explanation by example approach to show in detail the logic of the algorithm used to semantically compare the keywords extracted from the recommended tweet and the user interests .

\section{Empirical Study}

\subsection{Participants}
% recruitment: how long, where
% how many participants
% average time to complete the survey
% which were excluded
The target groups of our study were researchers and students who have at least one scientific
publication. Participants were recruited via e-mail, word-of-mouth, and groups in social media networks 
and had to fulfill two participation requirements: they had to have at least one scientific publication and a Semantic Scholar ID, which is required for the interest model inference step. To obtain a diverse sample, the study included participants from different countries, educational levels, and study backgrounds. A total of 36 participants completed the study. We ensured the data quality through the examination of redundant answering patterns (e.g.,  consistent selection of only one answering option) and attention checks (i.e., "Please answer ’disagree’ on this question"). Accordingly, five participants were excluded. The final sample consisted of \textit{N} = 31 participants (14 males, 17 females) with an average age of 32 years. Out of the 31 participants, 19 (61.3\%) reported to live in Germany, where 12 (38.7\%) were international users from eight different countries. All participants had sufficient English language skills to participate in the study. The highest level of education reported by most participants was \textit{master's degree} (61.3\%). The majority of participants (38.7\%) had a study background in \textit{Computer Science}. 17 participants (54.8\%) were considered as \textit{Twitter users} who reported to use Twitter at least 1 hour a week, where 14 (45.2\%) reported to never use Twitter in a typical week. 

\subsection{Study procedure}
While the study was originally planned as a laboratory experiment, due to the COVID-19 pandemic and its restrictions, we decided to conduct an online study. Each session was accompanied by a research assistant for technical support. The ethics motion to conduct the user study was approved by the Ethics Committee of the
Department of Computer Science and Applied Cognitive Science of the Faculty of Engineering
at the University of Duisburg-Essen on February 10, 2021. All participants gave informed consent to study participation. Participants first answered a questionnaire in SosciSurvey\footnote{https://www.soscisurvey.de} which asks for their Semantic Scholar ID and included questions about their demographics and personal characteristics. Next, participants were given a short demo video on how to use the RIMA application. Afterwards, participants were asked to (1) create an account using their Semantic Scholar ID, (2) explore the system and find matching recommendations to their interests, and (3) take a close look at each explanation provided by the system. After that, participants were asked to evaluate each of the six explanations in terms of seven explanation goals, namely transparency, scrutability, trust, effectiveness, persuasiveness, efficiency, and satisfaction \cite{tintarev2015explaining}. All participants evaluated the explanations in an iterative approach, by answering the same set of questions for each explanation. To avoid any order-related biases, the order in which participants rated the explanations was
randomized. They needed on average 48.09 minutes to complete the questionnaire (\textit{SD} = 9.40, range = 24.08-65.23). At the end of the session, participants were debriefed and compensated with the possibility to win one of five Amazon vouchers. 

\subsection{Measurements}
\label{sec:measurements}

\subsubsection{Personal Characteristics}
\label{sec:pc}
Our study included measurements of six personal characteristics, namely: need for cognition,
visualization familiarity, personal innovativeness, trust propensity, domain knowledge, and technical
expertise. For each personal characteristic, answers were given on a 5-point Likert scale, ranging from 1
("strongly disagree") to 5 ("strongly agree"). 
%We calculated scores of the measured personal characteristics as the average of the values reported for the corresponding items. 
%In addition, we captured participants’ demographiccharacteristics, including gender, age, educational level, country, and field of study.
Table \ref{tab:measurementsPC} shows the definitions and example items for each of the six measured personal characteristics. 

\textit{Need for Cognition:} Need for cognition (NFC) refers to the tendency for an individual to engage in and enjoy effortful cognitive activities \cite{cacioppo1984efficient}. To measure NFC, the NCS-6 by Lins de Holanda Coelho et al. \cite{lins2020very} was used, which is a short 6-item version of the original 18-item Need for Cognition Scale (NCS-18) by Cacioppo et al. \cite{cacioppo1984efficient}. By providing significant time savings, the NCS-6 benefits from reducing participant fatigue and enhancing the data quality for longer surveys \cite{lins2020very}. The reliability and validity of the NCS-6 are comparable to the original NCS-18, with an excellent internal consistency (Cronbach's $\alpha$ between .90 and .94). 
%Answers were given on a 5-point Likert scale, ranging from 1 (strongly disagree) to 5 (strongly agree). Example item: "I would prefer complex to simple problems"

\textit{Visualization Familiarity:} Visualization Familiarity (VF) refers to the extent to which users have experience with analyzing and graphing data visualizations. To measure VF, this study adopted the scale proposed in the work by Kouki et al. \cite{kouki2019personalized}. The internal consistency of this scale is excellent (Cronbach's $\alpha$ = .92). 
%Answers were given on a 5-point Likert scale, ranging from 1 (strongly disagree) to 5 (strongly agree).

\textit{Personal Innovativeness:} Personal innovativeness (PI) is a personality trait that represents an individual's confidence or optimism regarding adoption of new technologies or ideas. Depending on their degree of PI, individuals will be either more or less willing to adopt new technologies \cite{agarwal1998conceptual}. To measure PI, the scale by McKnight et al. \cite{mcknight2002developing} was used. The scale has good internal consistency (Cronbach's $\alpha$ = .89). 
%Answers were given on a 5-point Likert scale, ranging from 1 (strongly disagree) to 5 (strongly agree). Example item: "I like to explore new Web sites"

\textit{Trust Propensity:} Trust propensity (TP) describes the level of intensity of an individual's natural inclination to trust other parties in general \cite{komiak2003impact} and is related to the distribution to trust \cite{mcknight2002developing}. To measure TP, the scale by Lee and Turban \cite{lee2001trust} was used. The scale has excellent internal consistency (Cronbach's $\alpha$ = .90). 
%Answers were given on a 5-point Likert scale, ranging from 1 (strongly disagree) to 5 (strongly agree). Example item: "It is easy for me to trust a person/thing".

\textit{Domain Knowledge:} Domain knowledge (DK) refers to the users' knowledge about or experience with the type of recommended items (i.e., tweets). To measure DK, this study adopted the scale used in the work by Al-Natour et al. \cite{al2008effects}. The scale was adapted to the context of this study by changing the word "computer" to "Twitter". The internal consistency of this scale is excellent (Cronbach's $\alpha$ = .95). 
%Answers were given on a 5-point Likert scale, ranging from 1 (strongly disagree) to 5 (strongly agree). Example item: "I have extensive experience in Twitter"

\textit{Technical Expertise:} In this work, technical expertise (TE) refers to users' knowledge about artificial intelligence and recommender systems. To measure TE, this study adopted the scale used in the work by Kunkel et al. \cite{kunkel2021identifying}. 
%Example item: "I have expert knowledge about machine learning or artificial intelligence"
\begin{table}[!ht]
	\small
  	\centering
    \centering
    \begin{tabular}{|p{0.19\linewidth} | p{0.38\linewidth}| p{0.29\linewidth}| p{0.05\linewidth}|}

        \hline
        \textbf{Personal characteristics (PC)} & \textbf{Definition} & \textbf{Example item} & \textbf{Source} \\
        \hline
        Need for Cognition (NFC) &  Tendency for an individual to engage in and enjoy effortful cognitive activities \cite{cacioppo1984efficient} & I would prefer complex to simple problems. & \cite{lins2020very} \\
        \hline
        Visualization Familiarity (VF) & Extent to which users have experience with analyzing and graphing data visualizations & I frequently analyze data visualizations. & \cite{kouki2019personalized} \\
        \hline
        Personal Innovativeness (PI) & Confidence or optimism regarding adoption of new technologies \cite{agarwal1998conceptual} & I like to explore new Web sites. & \cite{mcknight2002developing}  \\
        \hline
        Trust Propensity (TP) & Level of intensity of an individual's natural inclination to trust other parties in general \cite{komiak2003impact} & It is easy for me to trust a person/thing. & \cite{lee2001trust} \\
        \hline
        Domain Knowledge (DK) & Knowledge about or experience with the type of recommended items & I am knowledgeable about Twitter. & \cite{al2008effects} \\
        \hline
        Technical Expertise (TE) & Knowledge about artificial intelligence and recommender systems & In the past I learned about how recommender systems work. & \cite{kunkel2021identifying} \\
        \hline
    \end{tabular}
    \caption{Measurement of personal characteristics.}
    \label{tab:measurementsPC}
\end{table}

\subsubsection{Explanation Goals}
\label{sec:exp_goals}
The measurements for the seven explanation goals were adopted from different previous works \cite{balog2020measuring,kouki2019personalized,tintarev2008effectiveness,tintarev2011designing,tintarev2012evaluating,wang2007recommendation,zhao2019users}. The first six explanation goals were measured using a 5-point Likert-scale, while satisfaction was measured using a 7-point Likert-scale. An overview of used questionnaire items is shown in Table \ref{table:explanationaimsmetrics}. Besides quantitatively measuring the explanation goals, participants could provide qualitative feedback to each explanation and the overall RS by answering a set of open-ended questions.

\begin{table}[!ht]
	\small
  	\centering
    \begin{tabular}{|l|l|l|}
    	\hline
    	\textbf{Metric} & \textbf{Statement} & \textbf{Source} \\
    	& This explanation ... & \\
    	\hline
    	Transparency & helps me to understand what the recommendations are based on. & \cite{balog2020measuring} \\
    	\hline
    	Scrutability & allows me to give feedback on how well my preferences have been understood. & \cite{balog2020measuring} \\
   		\hline
    	Trust (Competence) & shows me that the system has the expertise to understand my needs and preferences. & \cite{wang2007recommendation} \\
    	\hline
    	Trust (Benevolence) & shows me that the system keeps my interests in mind. & \cite{wang2007recommendation} \\
    	\hline
    	Trust (Integrity) & shows me that the system is honest. & \cite{wang2007recommendation} \\
    	\hline
    	Effectiveness & helps me to determine how well the recommendations match my interests. & \cite{tintarev2011designing} \\
    	\hline
    	Persuasiveness & is convincing. & \cite{kouki2019personalized} \\
    	\hline
		Efficiency & helps me to determine faster how well the recommendations match my interests. & \cite{tintarev2011designing} \\
    	\hline
    	& \textbf{Question} & \\
    	Satisfaction & How good do you think this explanation is? & \cite{tintarev2008effectiveness, tintarev2012evaluating} \\
    	\hline
    \end{tabular}
  \caption{An overview of questionnaire items used for the evaluation of explanations.}
  \label{table:explanationaimsmetrics}
\end{table}
\subsection{Study design}
The RIMA application explains the recommendations with three different levels of detail (basic, intermediate, advanced). Following a within-subjects design, participants rated the three explanations in terms of the seven explanation goals outlined above. We calculated scores of the measured personal characteristics as the average of the values reported for the corresponding items. Further, we calculated
the evaluation score for trust as the average of the individual values reported for the three
trusting beliefs (i.e., competence, benevolence, and integrity). 
\section{Results}
\subsection{Interaction effects}
To address our research question, we performed seven repeated measures ANCOVA analyses, where the evaluation scores of the seven explanation goals were included as dependent variables (DV), the explanation level (basic, intermediate, advanced) as independent variable (IV), and the personal characteristics scores as covariates. To stress here that we did not use ANCOVA to assess the overall effect of the IV (level of detail) on the DV (perception of explanation goal) while controlling for the covariate (personal characteristics), which would assume that there is no interaction between the IV and the covariate (i.e., homogeneity of regression slopes), but rather to find potential interactions between level of detail and personal characteristics. To visualize the significant interaction effects, we performed a median split for each personal characteristics dividing the participants in a low and high group for each of them.

\begin{figure}[h]
	\centering
	\begin{subfigure}[b]{0.3\textwidth}
		\includegraphics[height=0.6\textwidth, width=\textwidth]{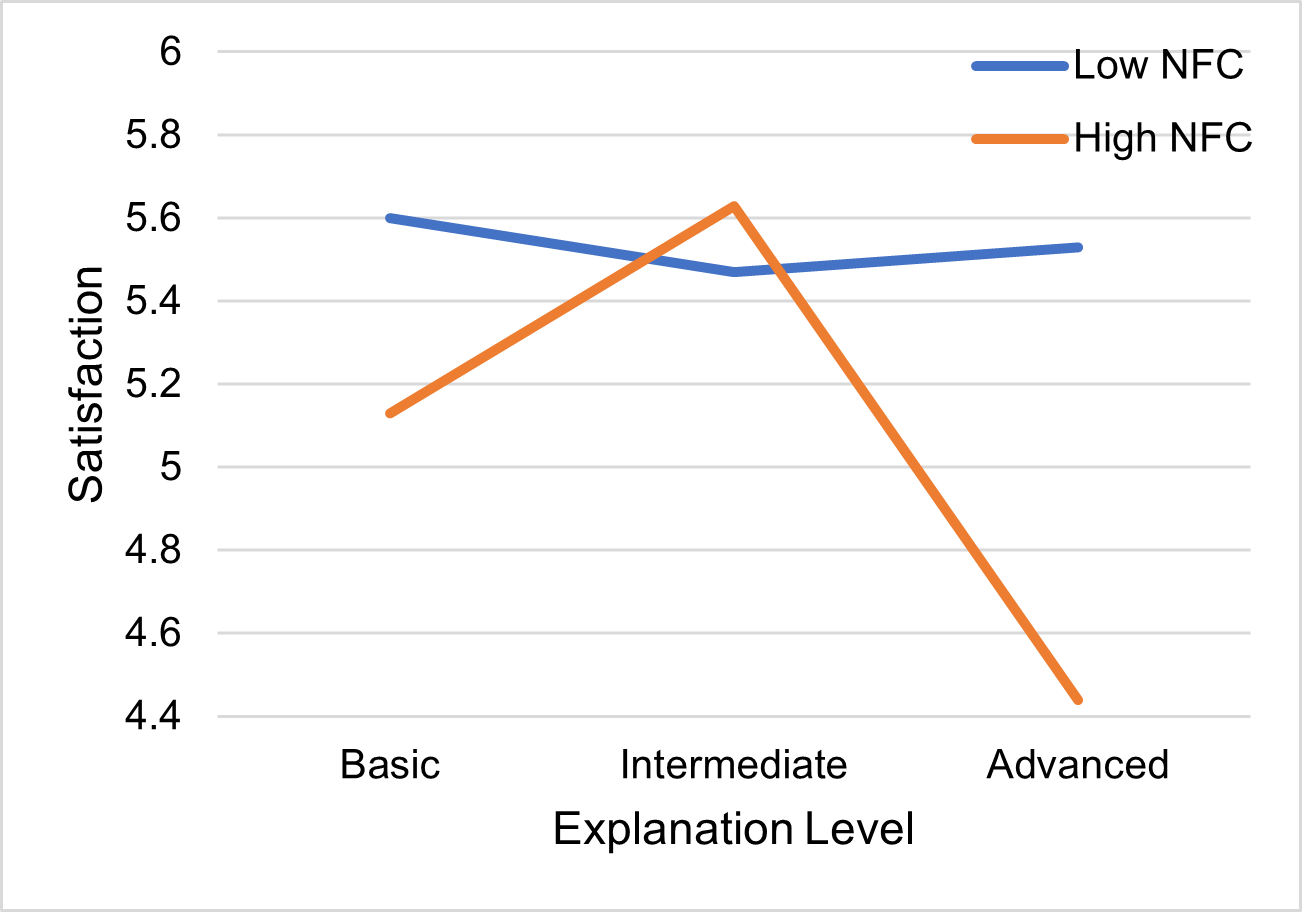}
		\caption{NFC and Satisfaction}
		\label{fig:rq2outputsatNFC}
	\end{subfigure}
\hfill
	\begin{subfigure}[b]{0.3\textwidth}
		\includegraphics[height=0.6\textwidth, width=\textwidth]{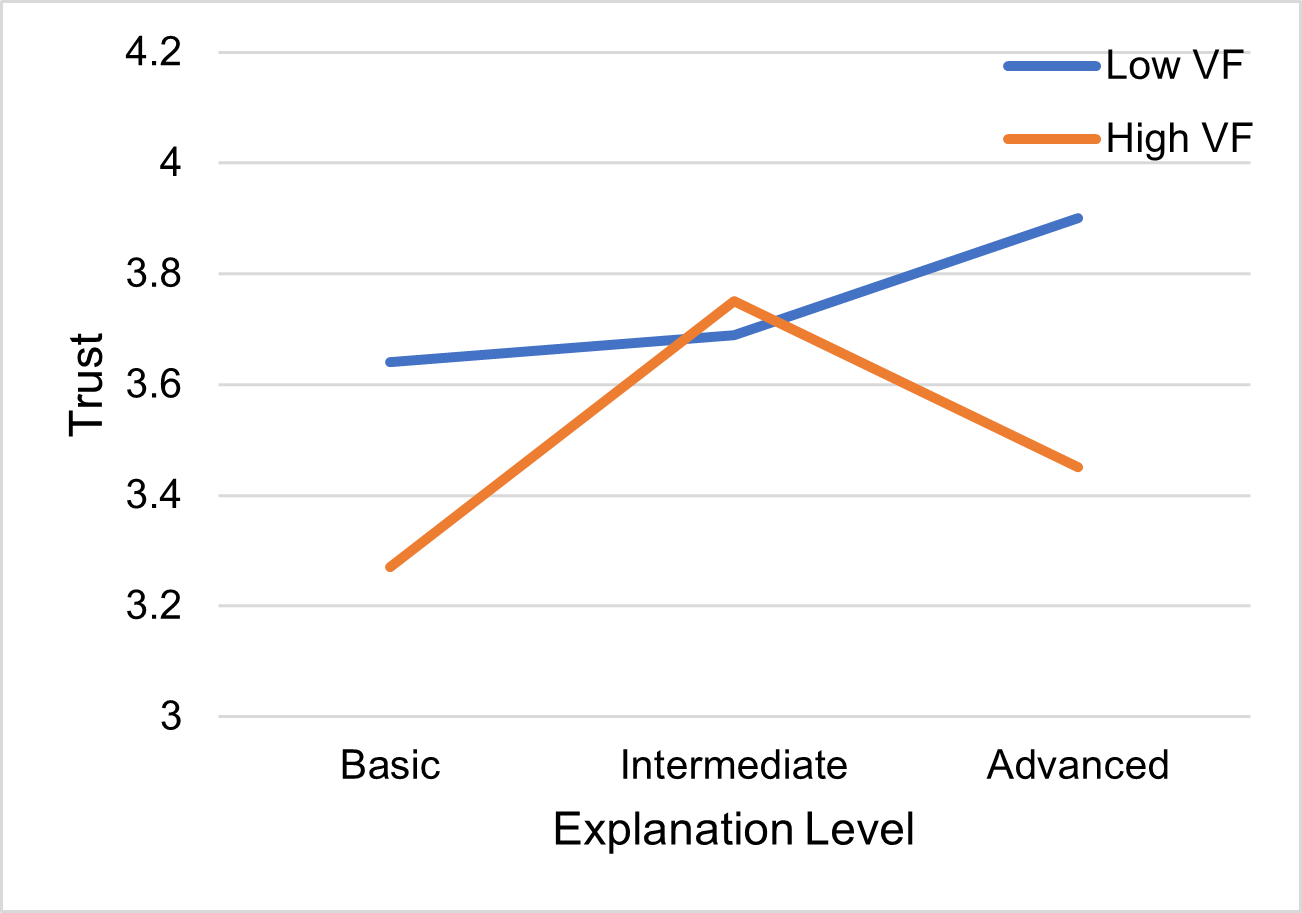}
		\caption{VF and Trust}
		\label{fig:rq2outputtrustVF}
	\end{subfigure}
\hfill
	\begin{subfigure}[b]{0.3\textwidth}
		\includegraphics[height=0.6\textwidth, width=\textwidth]{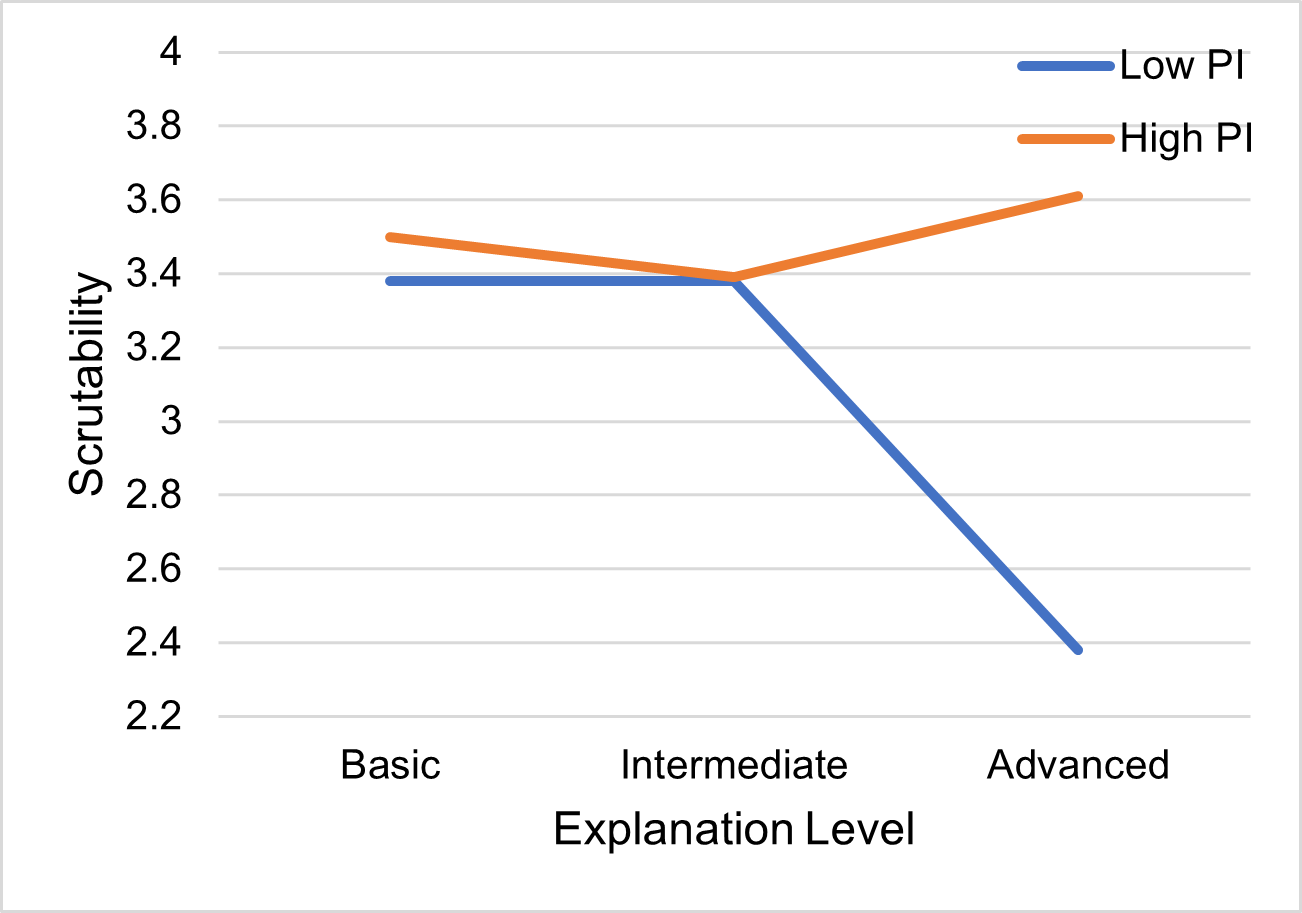}
		\caption{PI and Scrutability}
		\label{fig:rqoutputscruPI}
	\end{subfigure}
\hfill
	\begin{subfigure}[b]{0.3\textwidth}
		\includegraphics[height=0.6\textwidth, width=\textwidth]{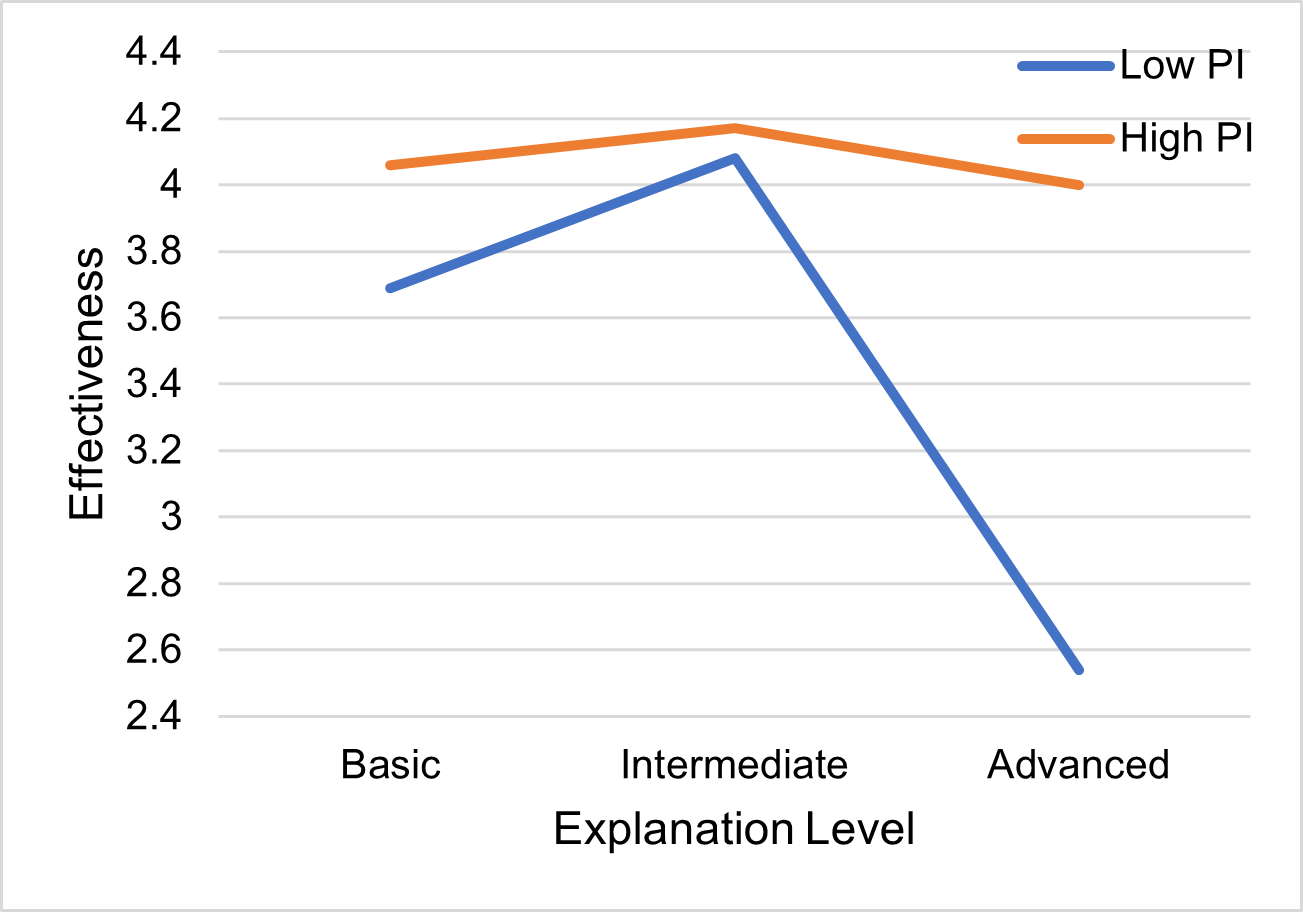}
		\caption{PI and Effectiveness}
		\label{fig:rq2outputeffecPI}
	\end{subfigure}
\hfill
	\begin{subfigure}[b]{0.3\textwidth}
		\includegraphics[height=0.6\textwidth, width=\textwidth]{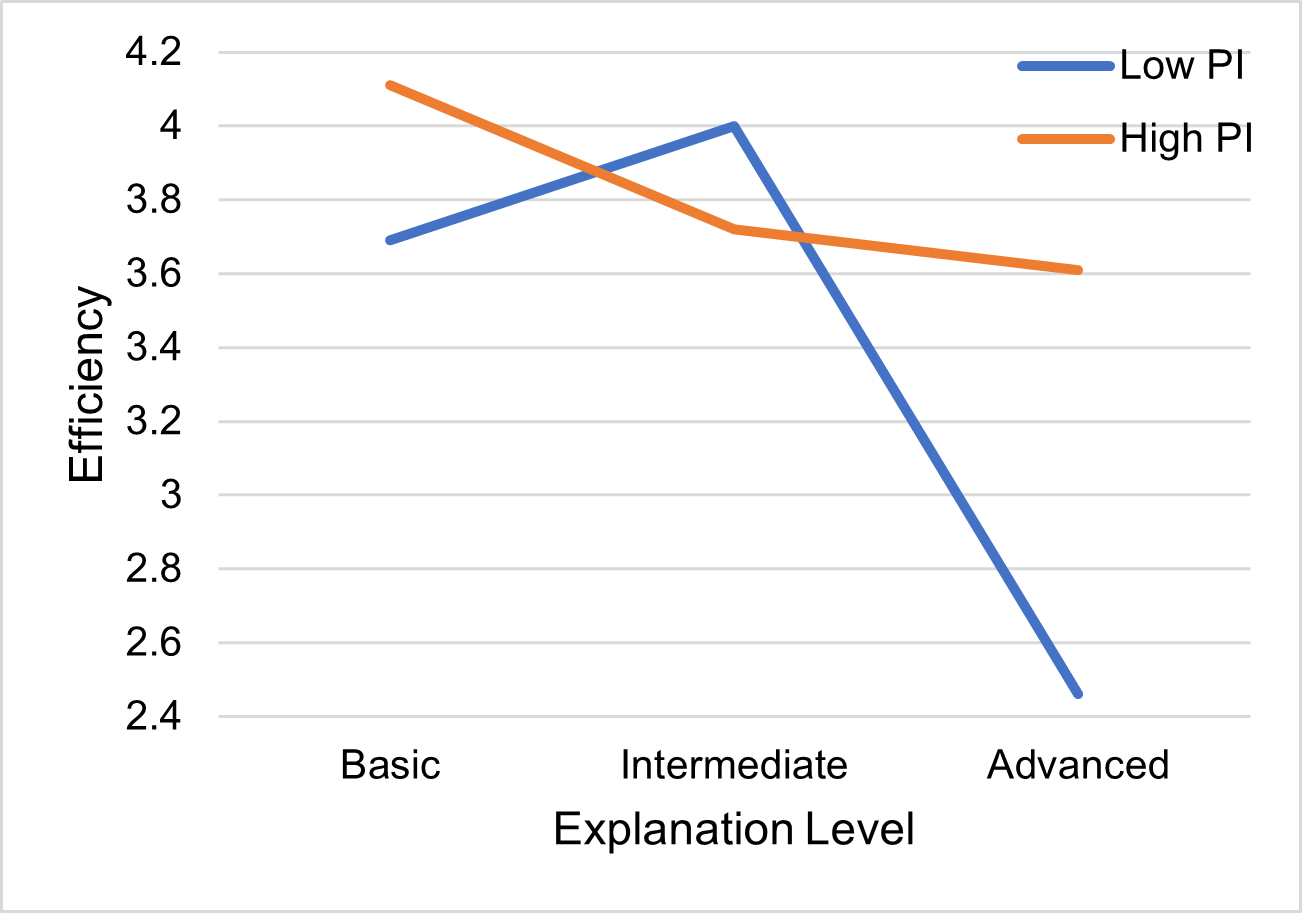}
		\caption{PI and Efficiency}
		\label{fig:rq2outputefficPI}
	\end{subfigure}
\hfill
	\begin{subfigure}[b]{0.3\textwidth}
		\includegraphics[height=0.6\textwidth, width=\textwidth]{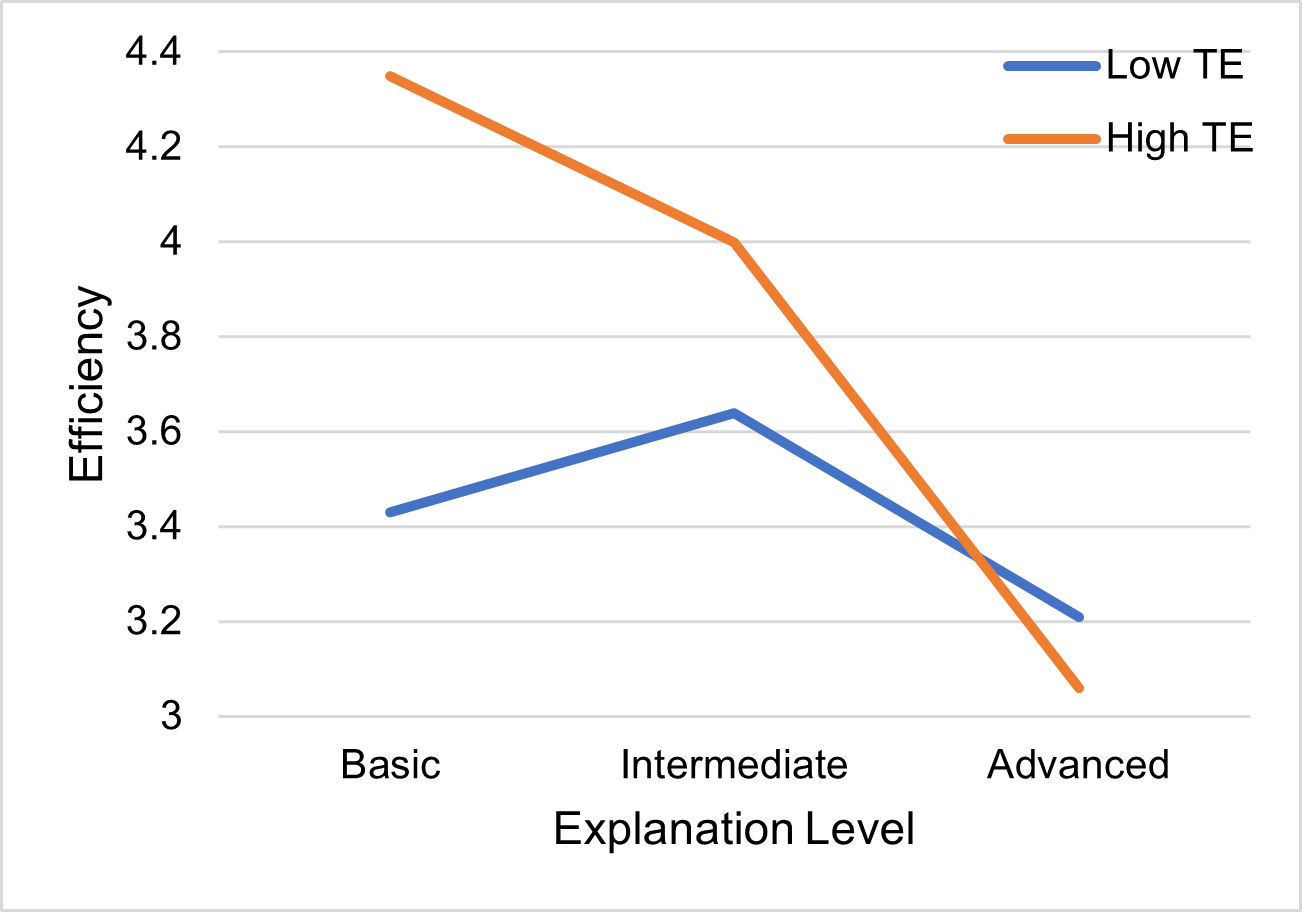}
		\caption{TE and Efficiency}
		\label{fig:rq2outputefficTE}
	\end{subfigure}
	\caption{The interaction effects between personal characteristics and explanation levels in terms of explanation goals.}
	\label{fig:rq2output}
\end{figure}

\textbf{Need for Cognition:} A significant interaction was found between NFC and explanation level in terms of \textit{satisfaction} (\textit{F}(2,48) = 3.654, \textit{p} = .033, \textit{f} = .39). The effect size corresponds to a moderate effect \cite{cohen1988statistical}. Figure \ref{fig:rq2outputsatNFC} shows that for users with high NFC, satisfaction first increased, while the advanced explanation had the lowest average satisfaction. No significant interactions between NFC and explanation level in terms of the other explanation goals were found.

\textbf{Visualization Familiarity:} A significant interaction was found between VF and explanation level in terms of \textit{trust} (\textit{F}(2,48) = 3.639, \textit{p} = .034, \textit{f} = .39). The effect size corresponds to a moderate effect \cite{cohen1988statistical}. Figure \ref{fig:rq2outputtrustVF} shows that, for users with low VF, trust increased with the explanation levels, while users with high VF had the average highest trust for the intermediate explanation. There were no significant interactions between VF and explanation level in terms of the other explanation goals.

\textbf{Personal Innovativeness:} A significant interaction was found between PI and explanation level in terms of \textit{scrutability} (\textit{F}(2,48) = 3.478, \textit{p} = .039, \textit{f} = .38), \textit{effectiveness} (\textit{F}(2,48) = 4.231, \textit{p} = .030, \textit{f} = .42), and \textit{efficiency} (\textit{F}(2,48) = 3.237, \textit{p} = .048, \textit{f} = .37). The effects are moderate for scrutability and efficiency, and strong for effectiveness \cite{cohen1988statistical}. The interaction plots in Figure \ref{fig:rqoutputscruPI}, \ref{fig:rq2outputeffecPI}, and \ref{fig:rq2outputefficPI} show that users with low PI had lowest average perception of scrutability, effectiveness and efficiency for the advanced explanation. There were no significant interactions between PI and explanation level in terms of transparency, trust, persuasiveness, or satisfaction.

%\textbf{Trust Propensity:} No significant interactions between TP and explanation level in terms of the seven explanation goals were found.

%\textbf{Domain Knowledge:} No significant interactions between DK and explanation level in terms of the seven explanation goals were found.

\textbf{Technical Expertise:} A significant interaction was found between TE and explanation level in terms of \textit{efficiency} (\textit{F}(2,48) = 3.262, \textit{p} = .047, \textit{f} = .37). The effect size corresponds to a moderate effect \cite{cohen1988statistical}. Figure \ref{fig:rq2outputefficTE} shows that users with high TE had higher average perceptions of efficiency for the basic and intermediate explanation than users with low TE.  No significant interactions between TE and explanation level in terms of the other explanation goals were found.

Finally, there were no significant interactions between both \textbf{Trust Propensity} and \textbf{Domain Knowledge} and explanation level in terms of the seven explanation goals.

\subsection{Qualitative analysis}
Besides our quantitative analysis, we also conducted a qualitative analysis of the open-ended questions to gain further insights into the reasons behind the individual differences in the perception of explanations. We followed the instruction proposed by \citet{braun2006using}. To do so, we started by familiarizing ourselves with the depth and breadth of the qualitative data. Next, we worked systematically through the data set and coded each answer to identify patterns in the data set. Then, we organized the codes into meaningful groups. The analysis was rather deductive as we aimed to find additional explanations for the findings of our quantitative analysis. 
%Moreover, codes regarding personalized explanations were inductive, i.e., fully data-driven.

\subsubsection{Explanations with different levels of detail}
One major theme reported by 26 participants in the open-ended questions was related to the general feature of providing explanations with different  levels of detail. Particularly, the majority of participants (23 of 26) reported that it is helpful to choose \textit{"which amount of information is enough for the respective recommendation”} (P24) and that it \textit{"gives the user the opportunity to learn/discover as much or as little as they want”} (P12). Nine participants agreed that the advanced explanation \textit{“might by a bit too much for a normal user”} (P17). Therefore, participants argued that it is better to provide different levels of detail to \textit{“satisfy the needs of different people”} (P2) so that \textit{“everyone can handle the system”} (P4), instead of  \textit{“providing all info in a single step”} (P5). Besides individual differences, a number of participants believed that the required amount of information will also depend on the specific context and situation: users will choose the required level of detail \textit{“based on one's curiosity”} (P7) and depending on \textit{"what is helpful at the moment”} (P3) or \textit{“when something goes wrong, I might be interested in more detailed explanations to fully understand the mistakes”} (P1). Only five participants reported that they do not wish to see explanations with different levels of detail, as they would only need one explanation \textit{"that the system thinks is "right" for me"} (P21) instead of offering \textit{“so many on the plate”} (P16).

\subsubsection{Need for Cognition}
% high NFC -> lower satisfaction with the advanced explanation (output)
One surprising finding of the quantitative analysis is that the overall average satisfaction with the advanced explanation of the recommendations was significantly lower for users with high than low NFC. 
%We expected that users with high NFC would be more satisfied with the detailed advanced explanation (as is the case for the recommendation input). 
After analyzing their answers, we observed that out of the comments criticizing the advanced explanation of the recommendations, the majority (12 of 15) came from users with high NFC. For instance, P1 reported \textit{“I was not able to fully read the explanation because some letters were too small”} and P2 reported \textit{“I recommend optimizing the view and the formatting of the figure”}. Further, three participants with high NFC disliked that the advanced explanation is \textit{“static”} (P22) and {“just about the algorithm”} (P26), so it \textit{“does not differ from tweet to tweet”} (P9). Overall, it seems that users with high NFC wanted to explore the explanation in detail, but were disappointed when they realized that it shows example values which were hardly readable, while users with low NFC had no need to read every small detail.

\subsubsection{Visualization Familiarity}

Similarly, the quantitative analysis revealed that users with high VF perceived the advanced explanation of the recommendations as less trustworthy than users with low VF. The qualitative analysis revealed that users with high VF disliked the static appearance of the advanced explanation of the recommendations. For instance, P18 reported \textit{“it looks like a standard explanation of the system”} and P28 reported \textit{“they do not show actual formulas (i.e. vector model) or complete data (length of feature vector and values)”}. In addition, one participant with high VF perceived the advanced explanation to be \textit{“a bit hidden"} and that it \textit{"should be made more prominent”} (P21). One possible explanation for the reduced trust is that users with more knowledge about data visualization spent more time to explore the explanation in detail and understand the interplay between the recommendation input and output. They  
might have imagined how the chart could look like if it showed the values of their actual feature vectors. As their expectations could not be met, this might have created beliefs that the system is not honest about how it calculates the similarity scores of their own recommendations, thus they had lower perceptions of trust. This assumption, however, requires further investigation.

\subsubsection{Personal Innovativeness}

% high PI -> lower perception of transparency of the intermediate explanation (input)
%We found that users with high PI perceived the intermediate explanation of the interest model to be the least transparent. After analyzing their answers, we observed that a number of participants with high PI disliked the general appearance of the explanation and suggested that \textit{“this one needs better UX”} (P16). As a high PI is related to higher interest in testing new websites, these users might have paid more attention to the general design and usability of the explanations and spent less effort in trying to understand how the system works. %(i.e. lower perceived transparency). 
%However, we also observed that a number of participants with high PI encountered technical issues with the extraction of their publications: P6 reported \textit{“my indexed study did not show the true study which I have done, it seems to be inaccurate”}. Overall, we assume that users with high PI had lower perceptions of transparency of the intermediate explanation as they detected design and technical issues.
%and P15 suggested to improve the \textit{“weighting system”}. 

% low PI -> lower scrutability, trust, effectiveness, efficiency (advanced output)
Another pattern that we identified in our quantitative analysis is that users with low PI perceived the advanced explanation of the recommendations to be least scrutable, effective, and efficient, and their perceptions of the advanced explanation were much lower than those of users with high PI. We observed that the majority (10 of 13) of participants with low PI reported to prefer the intermediate explanation as it is relates to their actual interests, which \textit{"leads to a quick validation of crossed interests"} (P5). One participant with low PI perceived the intermediate explanation to be more scrutable than the advanced explanation: \textit{“Intermediate one was what really helped me to understand why a strange tweet was recommended”} and \textit{“the advanced one in the tweets view was the less useful”} (P23). This seems to indicate that an explanation via example reduces their perceptions of scrutability, effectiveness, and efficiency.
%
%\subsubsection{Trust Propensity}
%
%%% low TP -> highest perception of scrutability in the basic explanation (input)
%We found that users with low TP had the highest perception of scrutability of the basic explanation of the interest model, while the intermediate and advanced explanation reduced their perceived scrutability. The qualitative analysis revealed that half of participants with low TP reported to prefer the basic explanation as it is \textit{“more fun and interactive”} (P12). One participant with low PI also reported to be familiar with this interaction component: \textit{“It's simple and I get used to other tools similar to this one”} (P16). As the intermediate explanation shows the weight of an interest, one participant with low TP reported \textit{“I do not know how the number is calculated”} (P6). Thus, as the basic explanation is easy to use and understand, it might have improved users' perception of scrutability.

\subsubsection{Technical Expertise}
% high TE -> higher perception of efficiency in the basic and intermediate explanation (output)
We found that users with high TE had higher perceptions of efficiency of the basic and intermediate explanation of the recommendations than users with low TE. After comparing their answers, we observed that users with high TE indeed perceived the basic and intermediate explanation as efficient: \textit{“The similarity score is sufficient for a quick review of the recommended tweets”} (P20) and \textit{“the Intermediate gives useful data quickly and easily”} (P12). In contrast, users with low TE seemed to have more difficulties in understanding the basic and intermediate explanation: \textit{“Found it difficult to fully understand the visualization”} (P4)” and \textit{“It's a little overwhelming at first”} (P18). 

\section{Discussion}
% summarize the results of the ANOVA analyses
% which are in line with research? / which were expected? -> citations
% which were surprising? -> we analysed the open-ended questions
% the answers suggest/seem to indicate that ...
% we believe ...

%In this work, we evaluated an explainable recommender system (RIMA) that explains both the recommendation input (i.e. user model) and recommendation output (i.e. recommended items) with three different levels-of-detail (basic, intermediate, advanced). We conducted an online-study using a within-subject design and applied a mixed-methods approach, that is, we used a combination of quantitative and qualitative evaluation methods. In the study, users evaluated each explanation in terms of seven explanation goals, namely: transparency, scrutability, trust, effectiveness, efficiency, persuasiveness, and satisfaction. Furthermore, users were presented with a number of open-ended questions to capture a broad range of feedback to the explanations and the overall system. We investigated the effects of explanation level and scope as well as personal characteristics on the perception of explanations. 

We discuss the findings of our study in relation to our research question: "How do personal characteristics impact user perceptions of the explanation level of detail in terms of different explanation goals?". Our results show that there is no universal rule such as “not too little and not too much” \cite{kizilcec2016much} to be applied when providing explanations with different levels of detail. It depends. The perception of explainable RS with different levels of detail is affected to different degrees by the explanation goal and user type. These effects are summarized in Table \ref{tab:relations} and discussed below.  

\subsection{Main findings}
% summarize for each PC what we found (patterns!)
% what still remains unclear? -> future work

%For the second research question, we included personal characteristics into the analysis in order to investigate individual differences in the perception of the explanations. To do so, we performed a set of 7 repeated measured ANCOVA analyses, once for the recommendation input and the recommendation output. The analyses revealed that personal characteristics have an influence on users' satisfaction and trust as well as their perceptions of efficiency, effectiveness, and scrutability.

% satisfaction
\textit{Satisfaction.} One main finding is that NFC influenced satisfaction with the explanations of the interest model. Users with a low NFC were more satisfied with the basic explanation. In contrast to the findings in \cite{cacioppo1984efficient} indicating that users with high NFC preferred the highly detailed advanced explanation,    
%Users with a low NFC were more satisfied with the basic explanation, while users with high NFC preferred the highly detailed advanced explanation. This is in line with the findings in \cite{cacioppo1984efficient}. In addition, 
our analysis revealed that users with high NFC were dissatisfied with the advanced explanation of the tweet recommendations as they expected it to explain the reasoning behind the recommendations using their actual interest and tweet keywords instead of example values. Thus, the explanation could not meet their increased need to understand the recommendation. Overall, the results indicate that users with high NFC are more satisfied with detailed explanations that are personalized to their own data.

% trust
\textit{Trust.} Similarly, we found that the advanced explanation of the recommendations led to reduced trust of users with high VF. The qualitative analysis showed that these users were disappointed by the static appearance of the advanced explanation. Moreover, our analysis indicated that they also needed the explanation to be more visible. Thus, as the advanced explanation was the one that required most interaction steps to see, this might as well has created beliefs that the system is not honest about its inner logic which further reduced its trustworthiness. Overall, our finding confirms that, besides the level of detail, other aspects such as the general design or usability also contribute to the perception of explanations.

% efficiency
\textit{Efficiency.} Another finding is that users with low TE perceived the basic and intermediate explanation of the recommendations as less efficient than users with prior knowledge about RS. We believe that users who already know how RS work needed less time to understand the explanations (e.g. similarity score and heatmap), thus they could determine faster how well a recommendation matches their interests. In contrast, users with low TE might have needed more time to understand the explanations, thus perceived them as less efficient. This shows that users' expertise is an important influencing factor that should be considered when designing explanations to improve efficiency.

% scrutability
\textit{Scrutability.} 
%Our analysis also indicated that the basic explanation of the interest model led to higher perception of scrutability for users with low TP. The answers of participants with low TP indicate that these users perceived the basic explanation as more scrutable because they already know this kind of interaction component from other websites. As distrusting users may have doubt against how the system processes their data, the other two explanations could have been unfamiliar to them and created feelings of uncertainty about how the system understands their interests. Further, we believe that detailed information about the algorithm or wrong assumptions made by the system have a greater negative impact for these users. For instance, if users with low TP cannot understand why the system calculated a wrong weight for a specific interest, they might not be able to estimate if the system correctly understands their interests. 
%Lastly, 
Our analysis also indicated that the intermediate explanation of the recommendations (i.e., heatmap) improved perceptions of scrutability for users with low PI, as this explanation helped them to validate the matching between their interests and the tweet recommendation. Thus, it seems that users with low PI need the explanation to be tailored to their actual data to give feedback about how well a recommendation relates to their interest.

%\begin{table}[!ht]
%	\small
%  	\centering
%    \begin{tabular}{|l|l|l|l|l|l}
%        \hline
%        \textbf{Goal} & \textbf{PC} & \textbf{Level of detail} & \textbf{Scope} & \textbf{Like (+) / Dislike (-)} \\
%        \hline
%        Satisfaction & low NFC & Basic & Input, Output & simple (+), easy to understand (+) \\
%        \hline
%        Satisfaction & high NFC & Advanced  & Input & detailed (+), static (-) \\
%        \hline
%        Satisfaction & high NFC & Advanced & Output & design (-), static (-) \\
%        \hline
%        Trust & high VF & Intermediate & Output & relates to profile (+) \\
%        \hline
%        Trust & high VF & Advanced & Output & static (-), hidden (-) \\
%        \hline
%        Efficiency & high TE & Basic & Output & quick overview (+), easy to understand (+) \\
%        \hline
%        Efficiency & low TE & Basic & Output & difficult to understand (-), overwhelming (-) \\
%        \hline
%        Scrutability & low TP & Basic & Input & familiar (+) \\
%        \hline
%        Scrutability & low PI & Intermediate & Output & relates to profile (+) \\
%        \hline
%    \end{tabular}
%    \caption{Relationships between goal, personal characteristics (PC), and level of detail  }
%    \label{tab:relations}
%\end{table}

\begin{table}[!ht]
	\small
  	\centering
    \begin{tabular}{|l|l|l|l|l|l}
        \hline
        \textbf{Goal} & \textbf{PC} & \textbf{Level of detail} & \textbf{Like (+) / Dislike (-)} \\
        \hline
        Satisfaction & low NFC & Basic & simple (+), easy to understand (+) \\
        \hline
        Satisfaction & high NFC & Advanced & design (-), static (-) \\
        \hline
        Trust & high VF & Intermediate & relates to profile (+) \\
        \hline
        Trust & high VF & Advanced & static (-), hidden (-) \\
        \hline
        Efficiency & high TE & Basic, Intermediate & quick overview (+), easy to understand (+) \\
        \hline
        Efficiency & low TE & Basic, Intermediate & difficult to understand (-), overwhelming (-) \\
        \hline
        Scrutability & low PI & Intermediate & relates to profile (+) \\
        \hline
    \end{tabular}
    \caption{Relationships between goal, personal characteristics (PC), and level of detail  }
    \label{tab:relations}
\end{table}

\subsection{Design guidelines}
To summarize the insights gathered through our study, we have compiled some suggestions for the effective design of explanations in RS.

\subsubsection{Explanation with different levels of detail}
In general, our work has confirmed that different users have different needs for explanation and require different explanation
levels of detail. Thus, it is important that explanations should be tailored to the user type and that RS should provide explanations with \textit{different levels of detail} to meet the demands of different users, as also suggested in \cite{millecamp2019}. Further, our analysis also showed that, besides user characteristics, other aspects, such as the specific context and situation, influence how much detail a user wants to see. For instance,  when users detect wrong assumptions made
by the system or receive unexpected recommendations, they may be interested in seeing more
details to understand why the system came to its decision. This supports previous findings that
users need explanations for a variety of reasons (e.g., curiosity or system errors \cite{putnam2019exploring}). Similarly, \citet{gedikli2014} found that users have a higher need to understand a recommendation when
the recommendation is questionable or unexpected.

\subsubsection{Goal-oriented, human-centered explanation design}
As different design choices such as explanation style, scope, format, or level of detail can be affected by the explanation goal and user type, it is crucial to follow a \textit{goal-oriented}, \textit{human-centered} approach to explanation design that starts with an understanding of the users’ goals and personal characteristics and then work backward to design explanations that best meet these goals and personal characteristics. To get at this, we need to provide mappings related to “which design choice instance is \textit{good} for which explanation goal?” (mapping: Goal $\,\to\,$ Design choice) and “which design choice instance is \textit{good} for which user type” (mapping: User $\,\to\,$ Design choice). As in general there is an interaction effect of design choice and user type on the perception of the explanation goal, we need to explore triplets of the form \textit{(G, U, D)} to illustrate these interaction effects, where \textit{G} is the explanation goal (e.g., transparency, scrutability, trust, effectiveness, persuasiveness, efficiency, satisfaction), \textit{U} is a pair representing a personal characteristic of a user, e.g., (need for cognition, high) or (visualization familiarity, low), and \textit{D} is a pair representing an instance of a design choice, e.g., (level of detail, intermediate) or (explanation format, textual). The (G, U, D) triplet can be interpreted as an association rule G, U $\,\to\,$ D, e.g., \textit{satisfaction, (need for cognition, low) $\,\to\,$ (level of detail, basic)} to express that “in order to achieve higher satisfaction, provide explanation with an advanced level of detail to users with high need for cognition". In our study, we found few such association rules (see Table \ref{tab:relations}). We encourage our fellow researchers to conduct more user studies to evaluate explanations
designed for different explanation goals and user types, and to use (G, U, D) triplets in order to formally model the relationships they might find in their studies between the explanation goals, personal characteristics, and design choices, in different application domains. 

% satisfaction, (need for cognition: low) -> (level of detail: low)
% satisfaction, (need for cognition: high) -> (level of detail: high)
% efficiency, (visualization familiarity, high) -> (level of detail: low)
% efficiency, (technical expertise, low) -> (level-of-detail: low)
% scrutability, (trust propensity, low) -> (level of detail: low)

\subsubsection{Personalized explanation} It is essential to
provide explanations by following a \textit{personalization-by-design} approach to tailor the explanations to the user’s context, i.e., goals and personal characteristics, as also  pointed out by \citet{ain2022framework} who recently proposed a multi-dimensional conceptualization framework for personalized explanations in RS. Personalization should not only happen at the design choice level (i.e., tailor the explanation style, scope, format, or level of detail) but also at the content level (i.e., tailor the explanation’s content to user data).  We can think of two strategies for promoting \textit{personalized explanation} in RS. The first one is to support \textit{manual personalization} by providing tools that enable users to control and tailor the explanations based on their needs and preferences (user-driven personalized explanation). For example, provide different explanation levels of detail and then hand over control to the user to actively choose the level of detail that she wants to see. The second strategy is to build \textit{automatic personalization} controlled by the explainable RS (system-driven personalized explanation). The (G, U, D) triplets can be used as association rules to provide automatic personalized explanation taking into consideration individual user’s context.

\section{Limitations}
% measurements: self-report
% technical issues: tweets, extraction of interests, loading times
% sample size
Although immense cares have been put into the planning of the user study, this work has some limitations. Firstly, the small sample size of the study. Therefore, the results of the study
should be interpreted with caution and cannot be generalized. A larger sample would probably
have yielded more significant and reliable results. Unfortunately, due to the study requirements that a participant should have at least one scientific publication and a Semantic Scholar ID, it was not possible to find more participants in time. Secondly, the measurement of personal characteristics and the evaluation of the system were based solely on self-report. Even though the data was carefully collected, this method has some error influences, such as dishonesty due to social desirability, which can lead to bias. 
Thirdly, some explanation goals could have been measured using objective instead of subjective measurements. For instance, efficiency can also be measured using log-data such as the total interaction time.
Moreover, there were some technical issues that some participants have encountered during the study, which may have negatively influenced their perception of the RIMA application and the explanations.
%(e.g. high loading times of the interest extraction). 
Finally, the current design of the different levels of detail was mainly the result of brainstorming
sessions involving the authors and students from the local university. The results of this study could have been different if we had designed and presented other explanations (e.g., personalized advanced explanation instead of just explanation by example).

\section{Conclusion and Future Work}
In this paper, we aimed to shed light on an aspect that remains under-researched in the literature on explainable recommendation, namely the effects of personal characteristics and level of detail on the perception of explanations in a recommender system (RS). To this end, we developed and evaluated a transparent Recommendation and Interest Modeling Application (RIMA) that explains the recommendations with three different levels of detail (basic, intermediate, advanced). The results of our study demonstrated that the explanation design should foremost be tailored to user’s context, i.e. goals and personal characteristics. From our findings, we suggested some design guidelines to be considered when designing explanatory interfaces that align with user’s context. Our work has implications for theory on personalized explanation interfaces in RS. It demonstrates the interaction effects of personal characteristics and level of detail on the perception of explainable RS and it provides evidence for a dependency relation between explanation goal, user type, and design choice. Further, this work contributes to the practice by offering suggestions for the appropriate design of personalized explanation interfaces in RS. In future work we will assess the generalizability of the results in different application domains. We will also explore other possible visualizations to provide explanations at the three levels of detail. In particular, we will develop and evaluate advanced explanations that are tailored to user data. Further, we will investigate the interaction effects of personal characteristics and other design choices such as explanation style, scope, and format on the perception of explainable RS.

\bibliographystyle{ACM-Reference-Format}
\bibliography{imab-references}

\end{document}